%% file: newsembed_finetuning.tex
\def\arxiv{1}
\ifx\arxiv\undefined
    \documentclass[sigconf]{acmart}
\else
    \documentclass[sigconf,nonacm]{acmart}
\fi

\usepackage{makecell}
\usepackage[ruled,vlined]{algorithm2e}
\usepackage{caption}
\usepackage{subcaption}
\usepackage{natbib}
\usepackage[colorinlistoftodos]{todonotes}
\AtBeginDocument{%
  \providecommand\BibTeX{{%
    \normalfont B\kern-0.5em{\scshape i\kern-0.25em b}\kern-0.8em\TeX}}}

\usepackage{multirow}
\usepackage{array}
\newcolumntype{L}[1]{>{\arraybackslash}p{#1}}
\newtheorem{theorem}{Theorem}
\newtheorem{observation}[theorem]{Observation}

\copyrightyear{2021}
\acmYear{2021}
\setcopyright{rightsretained}
\acmConference[KDD '21]{Proceedings of the 27th ACM SIGKDD Conference on Knowledge Discovery and Data Mining}{August 14--18, 2021}{Virtual Event, Singapore}
\acmBooktitle{Proceedings of the 27th ACM SIGKDD Conference on Knowledge Discovery and Data Mining (KDD '21), August 14--18, 2021, Virtual Event, Singapore}\acmDOI{10.1145/3447548.3467392}
\acmISBN{978-1-4503-8332-5/21/08}
\begin{document}
\fancyhead{}
\title{NewsEmbed: Modeling News through Pre-trained Document Representations}


\author{Jialu Liu$^{*}$, Tianqi Liu$^{*}$, Cong Yu }
\affiliation{
  \institution{Google Research}
  \institution{\{jialu,tianqiliu,congyu\}@google.com}}

\thanks{\hspace{.3em} * Contribute to this work equally. Correspondence to: Jialu Liu.} 
  

\begin{abstract}
Effectively modeling text-rich fresh content such as news articles at document-level is a challenging problem. To ensure a content-based model generalize well to a broad range of applications, it is critical to have a training dataset that is large beyond the scale of human labels while achieving desired quality.
In this work, we address those two challenges by proposing a novel approach to mine semantically-relevant fresh documents, and their topic labels, with little human supervision. Meanwhile, we design a multitask model called NewsEmbed that alternatively trains a contrastive learning with a multi-label classification to derive a universal document encoder.
We show that the proposed approach can provide billions of high quality organic training examples and can be naturally extended to multilingual setting where texts in different languages are encoded in the same semantic space.
We experimentally demonstrate NewsEmbed's competitive performance across multiple natural language understanding tasks, both supervised and unsupervised.
\end{abstract}

\ifx\arxiv\undefined

    
\fi



\maketitle

\section{Introduction}
\label{sec: intro}
\input{src/introduction}

\section{Data Collection}
\label{sec: data_collection}
\input{src/data_collection}

\section{Models}
\label{sec: models}
\input{src/models}

\section{Evaluation}
\label{sec: evaluation}
\input{src/evaluation.tex}

\section{Related Work}
\label{sec: related_work}
\input{src/related_work}

\vspace{2mm}
\section{Conclusion and Future Work}
\label{sec: conclusions}
\input{src/conclusions.tex}

\newpage

\bibliography{newsembed_finetuning.bib}

\appendix
\input{src/appendix.tex}

\end{document}

%% file: src/introduction.tex
Effectively modeling text-rich news content on the public web can enhance a broad set of applications in information retrieval and recommendation, and is especially crucial for long standing news tasks such as story clustering and topic classification. It is, however, a challenging problem due to out-of-vocabulary concepts and evolving relations between entities.


A typical way to represent general text data is projecting it into a sparse or dense vector space. If the subsequent task is unsupervised, e.g., clustering, one can compute similarity between texts by applying predefined functions such as inner product. If the subsequent task requires additional training, e.g., classification, an additional task-specific fine-tuning can be applied. Our work follows this paradigm and specifically focuses on text-rich news content. We demonstrate the performance of our embedding model through comprehensive evaluations in both unsupervised and supervised setups.

Modeling text data has a rich history. Classical approaches involve statistical modelling of words to represent the whole document, such as bag-of-words \citep{joachims1998text}, term frequency-inverse document frequency \citep{kim2006automatically}, and topic modeling \citep{blei2003latent}. There is no notion of similarity between words in those systems and texts are represented by sparse word vectors. To address this, models such as Word2Vec \citep{mikolovdistributed}, GloVe \citep{pennington2014glove}, and FastText \citep{bojanowski2017enriching} have been developed to capture the co-occurrences of characters or words where there can be dense vector representations. With the development of deep learning, convolutional neural networks \citep{zhang2015text}, recurrent neural networks \citep{zhou2015c, kowsari2019text}, and transformer \citep{vaswani2017attention} are used to represent contextual information \citep{melamud2016context2vec, peters2018deep}. 
More recently, with the development of self-supervised, unsupervised, and transfer learning, pre-training techniques such as BERT \citep{devlin2019bert}, 
T5 \citep{raffel2020exploring}, and GPT3 \citep{brown2020language} are proposed to leverage the free text on the web and achieve the state-of-the-art performance on various natural language processing (NLP) tasks without the need for huge amount of labelled data.

Although above deep learning approaches show great advantage in learning \textit{token}-level contextual representations from self-supervision, they still require a good amount of labelled fine-tuning data to produce high quality \textit{document}-level embeddings, which are essential for news tasks such as clustering. Naive approaches have been shown to be even less competitive to averaged GloVe \citep{pennington2014glove} in terms of
semantic similarity \citep{reimers2019sentence}. The high cost of annotating supervised training data impedes the development of high quality models in most research or even industry settings in that direction. At \textit{sentence}-level, several studies have been leveraging existing large scale datasets. Universal Sentence Encoder (USE) \citep{cer2018universal} leverages open Web data sources such as Wikipedia and uses unsupervised learning augmented with additional training on Stanford Natural Language Inference (SNLI) corpus \citep{bowman2015large}. Sentence-BERT (SBERT) \citep{reimers2019sentence} finetunes siamese and triplet network structures on SNLI and the MultiGenre NLI dataset (MNLI) \citep{williams2018broad} to obtain competitive performance on Semantic Textual Similarity (STS) tasks \citep{cer2017semeval}. 

Both USE and SBERT are developed for sentence-level representation and it is questionable whether they can capture semantics given longer context. Moreover, while both show great potential by transfer learning from NLI to STS, it is still challenging to adapt them to news due to the many out-of-vocabulary concepts, entities, and events. Recent studies show that domain specific training data can usually improve tasks through domain adaptation \citep{ben2007analysis, gururangan2020don}. 
It is also important for a news content representation model to be cross-lingual so that semantically-similar texts in different languages can be aligned in the same vector space. Recent works \citep{conneau2020unsupervised, xue2020mt5} show that the knowledge can be transferred from high resource languages to low resource languages effectively by forcing alignments between different languages.

Inspired by those prior works, our goal is to leverage large scale open data in a smart and multi-lingual way to construct a document-level representation model specifically for the news domain, as part of our mission to understand news content better and improve a broad range of news applications.


We propose a system called NewsEmbed that trains an encoder from weakly supervised data. We first introduce an approach to obtain billions of cross-lingual document \textit{triplets} (anchor, positive, negative) where anchor and positive documents are semantically close, while anchor and negative documents are related but less similar. In addition, we describe an effective heuristic to mine large-scale document-topic associations with high precision. Regarding the modeling, we adopt triplet neural structure and co-train a BERT-initialized encoder from the previously mentioned two datasets, with InfoNCE loss \citep{oord2018representation} as the contrastive learning objective and BCE loss \citep{murphy2012machine} as the classification objective respectively.  Our contributions of this work are three-fold:
\begin{itemize}
\item We propose a scalable approach to obtain multilingual weak supervision data with the purpose of pre-training document representations in news domain. 
\item We develop an effective neural architecture to co-train contrastive learning with multi-label classification, which demonstrates strong performance in later unsupervised and supervised applications, some of which are even out-of-domain.
\item As a side product, we show that a BERT model pre-trained from news corpus alone performs competitive on generic cross-lingual benchmarks \citep{hu2020xtreme}. 
\end{itemize}

The rest of the paper is organized as follows. Section~\ref{sec: data_collection} will introduce our pipeline of collecting pre-training and weakly supervised datasets. Section~\ref{sec: models} will cover the model and learning approach. In Section~\ref{sec: evaluation}, we will compare our approach with other works and evaluate them with public benchmark datasets. Ablation studies will be provided to analyze alternative designs. In Section~\ref{sec: related_work}, related works will be discussed, followed by the conclusion in Section~\ref{sec: conclusions}.

%% file: src/data_collection.tex
NewsEmbed trains a transformer encoder initialized from a BERT pre-training checkpoint. In this section, we first introduce the pre-training dataset and then describe in details how to mine weak supervision for the next stage of training. 

\subsection{Corpus for Pre-training}
\label{subsec: corpus_pretraining}
We have two types of data: monolingual documents and
bilingual translation pairs fed to a BERT masked language model.

\subsubsection{Monolingual Documents}
We collect $\sim$1.5B newsy articles based on an in-house crawler that extracts structured information including title, body, author, byline date and anchors. Documents without author or byline date are filtered out. To ensure the document is text-rich, we discard any document with fewer than 100 words. Furthermore, we apply similar data cleaning heuristics in \citep{raffel2020exploring} to deduplicate the dataset and remove low-quality content. 

It is worth noting that there exist open sourced, easy-to-use crawlers, such as news-please\footnote{https://github.com/fhamborg/news-please}, that can parse and extract structured news articles. They recursively follow internal hyperlinks and RSS feeds to fetch both recent and also old, archived documents.

\subsubsection{Bilingual Translation Pairs}
To better align the cross-lingual embedding space and transfer knowledge from high resource languages to low resource languages, a translation corpus is constructed from the web pages using a bitext mining system described in \cite{feng2020language}. This corpus contains $\sim$15B translation sentence pairs.

\begin{figure}
    \includegraphics[width=0.4\textwidth]{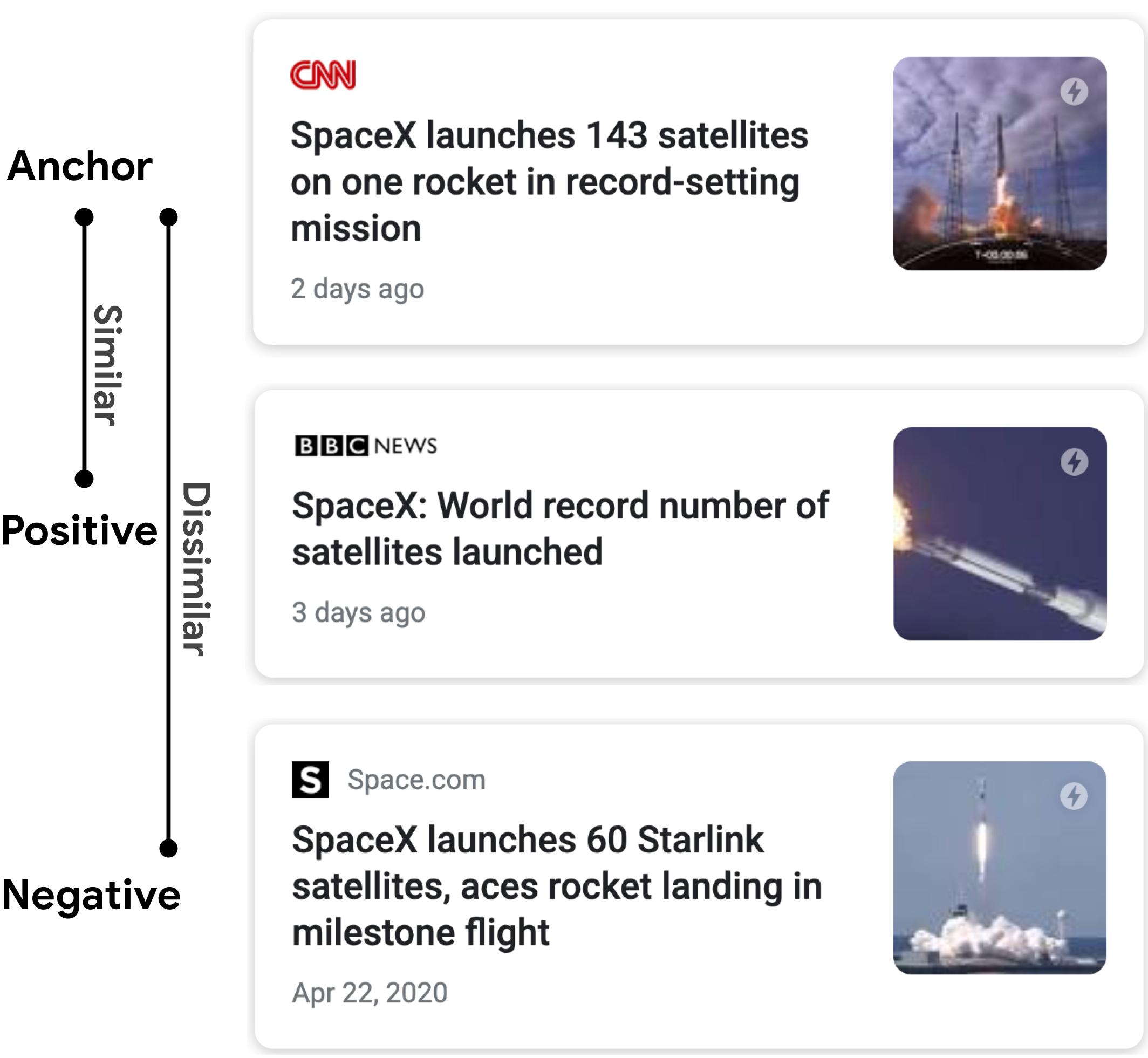}
    \caption{Example of a document triplet.}
    \label{fig:triplet}
    \vspace{-3mm}
\end{figure}

\begin{figure*}
    \includegraphics[width=0.8\textwidth]{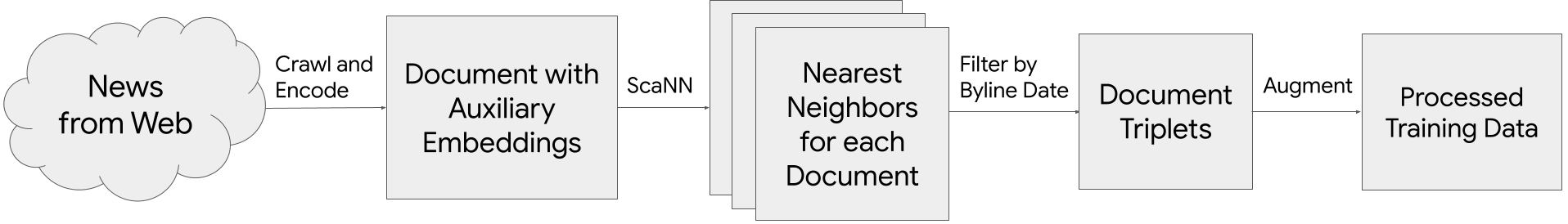}
    \caption{Data pipeline to collect document triplets for contrastive learning.}
    \label{fig: data_pipeline}
\end{figure*}

\subsection{Corpus for Weak Supervision}
Recall that the goal of this work is to train a \emph{document} encoder such that both in-domain supervised and unsupervised learning can benefit from it.
Typical pre-trained models, such as BERT variants, aren't effective without domain adaption including the modeling of both long context and different components in a document. This subsection provides details how we fill the gap by mining knowledge from the in-domain corpus.

\subsubsection{Document Triplets}\label{subsubsec: document_triplets} Inspired by contrastive learning initiated in computer vision \citep{le2020contrastive}, we believe text representations can be learned by enforcing similar elements to be equal and dissimilar elements to be different. When it comes to learning document representation, it naturally requires document triplets. Without loss of generality, we denote a triplet as (anchor, positive, negative), or ($a$, $p$, $n$) in short, where <$a$, $p$> document pair is more semantically similar than <$a$, $n$>. An example is illustrated in Figure~\ref{fig:triplet}.

Due to the special property of news, it is feasible to mine such document triplets based on the following two observations:
\begin{observation}(Redundancy)
News events or stories, especially the most popular ones, are likely to be reported and discussed from multiple publishers.
\end{observation}
\begin{observation}(Time-Sensitivity)\label{obs: time}
Lifetime of a news event or story is usually short while long running ones will evolve over time.
\end{observation}

The first observation ensures the abundance of organic <$a$, $p$> pairs if we have an effective approach to find these relevant documents. Since these texts are published by different publishers, they likely include paraphrasing of the same facts but expression of different opinions. From this perspective, by enforcing similarity between these documents, we are learning a semantic space where texts are close if they are targeting the same event or story. Also note that this source is different from typical text augmentation tricks, such as back translation and sentence reordering, because the paraphrasing is more natural and thus the quality is better.

The second, time-sensitivity, observation provides an effective signal to filter out noisy document triplets with high precision. To achieve this, assuming there exist some orthogonal signals to generate candidate triplets, we can apply aggressive filtering to ensure <$a$, $p$> is close in byline date while <$a$, $n$> is not.
We concede that such cleaning approach causes coverage loss, but empirically it leads to model improvement, implying that the quality of the weak supervision is more important than the quantity. 


The overall process for collecting document triplets is illustrated in Figure \ref{fig: data_pipeline}. We first encode each document using some auxiliary embedding signals such as entity and image embeddings. ScaNN \citep{avq_2020} is then used to extract approximate nearest neighbors for each document  w.r.t these auxiliary embeddings. Based on the nearest neighbors, we generate  ($a$, $p$, $n$) document triplets and apply the aforementioned byline date to improve the data quality. Lastly, these triplets are augmented to model both short- and long-context texts in the final dataset.
In the rest of this subsection, we discuss these steps in details.

\paragraph{Auxiliary embeddings for retrieving related documents}
In Figure~\ref{fig: data_pipeline}, the first and second steps are to extract semantically-related documents from the pre-training corpus in large scale. One can resort to various document embedding algorithms proposed in the literature.  And there is a high level of tolerance to the noise in these embedding methods because we have a denoising step later.

Concretely speaking, in NewsEmbed we compute three auxiliary embeddings for each document and use each of them to fetch related documents independently, including:

\begin{itemize}
    \item Entity embedding: Topical knowledge base entities are identified and their embeddings~\citep{zhang2016collaborative} are averaged to represent a document.
    \item Image embedding: Thumbnail embedding is derived using wavelet hashing~\citep{venkatesan2000robust}.
    \item Text embedding: Token embeddings in the title and keywords are averaged to represent a document~\citep{Chang2020Pre-training}. 
\end{itemize}

In particular, the former two embeddings are cross-lingual as their embedding sources are language-agnostic.

\paragraph{Triplet generation and filtering} After applying ScaNN \citep{avq_2020} to search nearest neighbors for each document in each auxiliary embedding space, we obtain large quantities of document pairs. The next step is transforming them to triplets by finding high quality <$a$, $p$> and <$a$, $n$> pairs. To achieve these, we merge all top-$K$ nearest neighbors for a document from all the auxiliary embedding types, and check byline date in the neighbor documents to decide whether the pair should be positive or negative. Empirically we set one day as maximum time difference for <$a$, $p$> and one year as minimum time difference for <$a$, $n$>. Since $a$ and $n$ are close in at least one of the auxiliary embedding space, we consider them as hard negatives so that later contrastive learning could be more effective.

Nevertheless, there is a shortcoming of heavily relying on byline date to denoise the triplets. We notice many of the false negatives from the above approach are similar in nature, i.e., document content is evergreen. For example, topics like ``minimum wage'' are mentioned frequently over time and thus are not time-sensitive as described in Observation~\ref{obs: time}. In this regard, we train an auxiliary \emph{byline date prediction model} specifically to predict byline date based on document title and body, with the motivation that the publication time of evergreen content is more difficult to predict compared to time-sensitive news events and stories. A logistic regression model is then deployed to remove false negatives by considering predicted date distribution together with the embedding similarities mentioned previously.
Since the number of features used here is very small, only a few hundred training examples are needed.

\paragraph{Augmented triplets}
After mining document triplets, we augment them in various ways as shown in Table~\ref{tab: triplet_source} before feeding the training data into the triplet neural network. The rational of the augmentation is to make the model learn a better embedding on both short context (i.e., title or anchor text) and long context (i.e., body alone or title concatenated with body). It is worth noting that augmented positive does not necessarily come from positive document, and we intentionally ensure that the augmented positive and negative are always of the same type,  to prevent biases of content length during the optimization of contrastive objective.

\begin{table}[]\small
\begin{tabular}{|r|r|r|}
\hline
Augmented Anchor          & Augmented Positive                 & Augmented Negative     \\\hline
Title $(a)$        & Body $(a)$        & Body $(n)$ \\
Anchor Text$^{*}$ $(a)$  & Title + Body $(a)$ & Title + Body  $(n)$\\
Title $(a)$      &  Title + Body $(p)$ &  Title + Body $(n)$ \\
Title + Body $(a)$      &  Title + Body $(p)$ &  Title + Body $(n)$\\\hline
\multicolumn{3}{L{7.8cm}}{$^{*}$\footnotesize{We consider sometimes the text of incoming anchor summarizes the document. To denoise, one can add a similarity check using the auxiliary text embedding.}} \\
\end{tabular}
\vspace{1mm}
\caption{Augmented triplets from document triplet $(a, p, n)$. Symbol + denotes segment concatenation.}
\label{tab: triplet_source}
\vspace{-7mm}
\end{table}

To encourage better cross-lingual alignment, for augmented triplets that are entirely not in English, we translate positive and negative into English. This helps the model to transfer knowledge in a multilingual setup.

\subsubsection{Document-Topic Associations}\label{subsec: document-topic}
Document triplets capture semantic similarity at the granularity of events or stories. To that extent, it lacks knowledge of coarse-grained topics to teach an encoder whether a document belongs to the category of ``investment'', ``lifestyle'' or some others. Adding this missing piece in the dataset can further enhance the representation power of the embedding.

\begin{figure}[t]
    \includegraphics[width=0.45\textwidth]{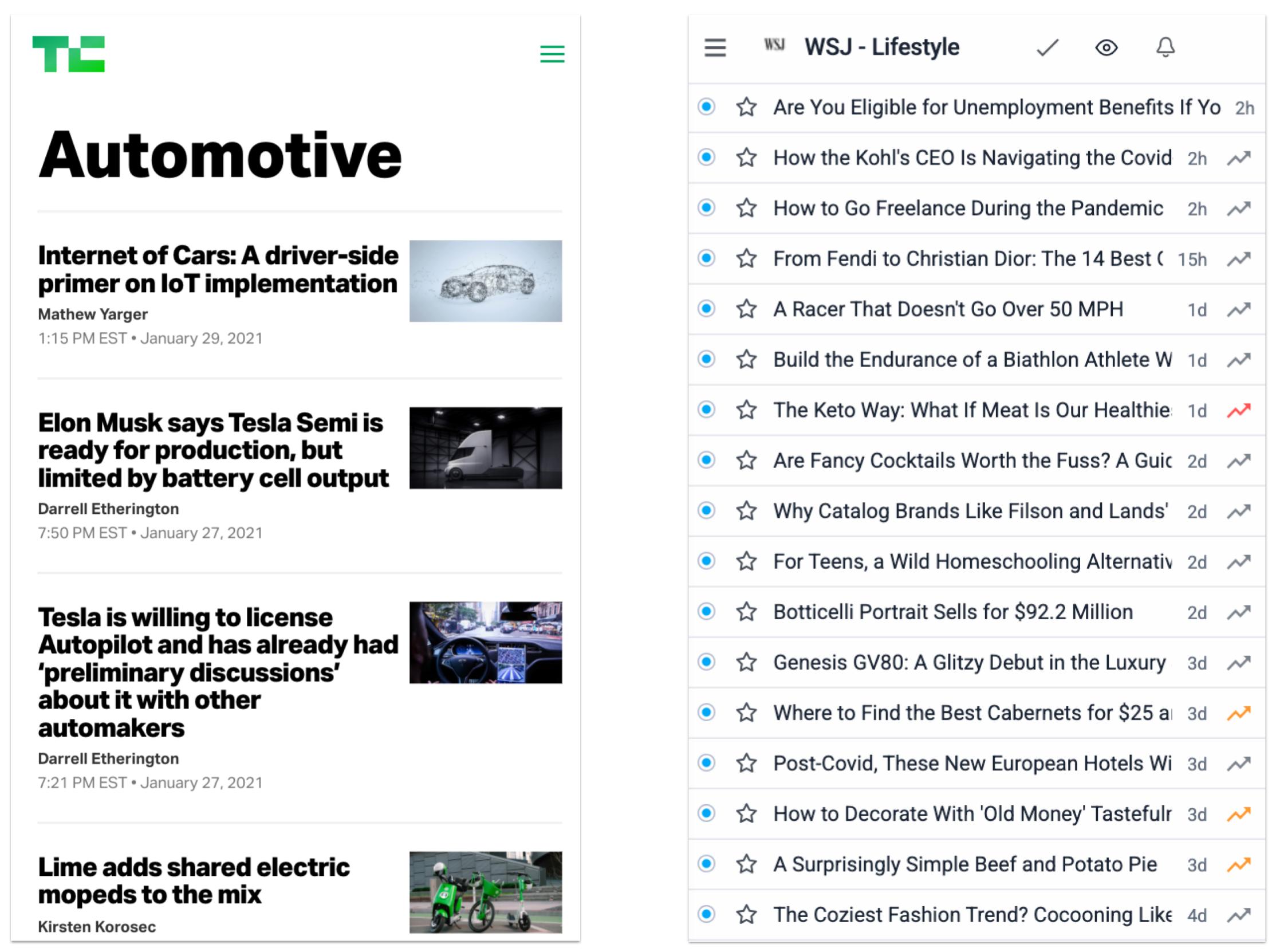}
    \caption{Left: A sub-directory page about automotive; Right: RSS feed for lifestyle.}
    \label{fig:hub_rss}
    \vspace{-4mm}
\end{figure}

To mine these document-topic associations in large-scale, we need to rely on an observation described as below:
\begin{observation}(Aggregation)
Publishers choose to aggregate documents with shared topics for ease of information consumption. 
\end{observation}
Typical examples of these ``hubs'' include sub-directories and RSS feeds of a site as shown in Figure~\ref{fig:hub_rss}.
However, not all ``hubs'' on the web are useful for our use case, e.g., many are based on author or publishing time.  Realizing this, we apply entity linking to hub page titles and derive a list of the most popular candidates. Then we select a subset that can be considered as coarse-grained topics and ignore the rest. In addition to regular broad topics, we notice some hub entities can be mapped to journalist types such as ``interview'' and ``opinion''. They are also included in our list to help learn the story styles.

A challenge in processing this dataset is that the mined document-topic associations only provide positive labels. In other words, it is not yet clear whether any unobserved <document, topic> pairs are negative. While different publishers tend to use different aggregation strategies consistently over time, we can treat unobserved <document, topic> pairs as negative if (and only if) that topic appear on the hub page of the same publisher. Additional sampling is applied to ensure positives and negatives are balanced.

%% file: src/models.tex
In this section, we present the model learning details in pre-training and weak supervision phases respectively.

\subsection{Language Model Pre-training}
\label{subsec: pretrain}
\input{src/pre-training.tex}

\subsection{Multitask Weakly Supervised Learning}
\label{subsec: train}
\input{src/training.tex}

%% file: src/pre-training.tex
Following the multilingual pre-training recipe, we train a BERT-style transformer encoder by Masked Language Modeling \citep{devlin2019bert} and Translation Language Modeling \citep{lample2019cross} on whole word masking \citep{cui2019pre}. Considering the page limit and novelty of this part, we omit some of the model learning details. Interested readers can refer to existing work for the language model formulation.

One difference from the previous work is that our translation sentence pairs are much shorter than monolingual documents. In this regard, we greedily pack multiple pairs into one sequence and reset position ids at each pair's head. Special treatment on self-attention layer is needed to prevent cross-attention among packed texts. Our solution is to correct the attention matrix by masking out cross-pair entries.


%% file: src/training.tex
Recall that we have two datasets collected as weak supervision: augmented document triplets and document-topic associations. At a high level, we treat them as training sources for two separate tasks with different loss functions, but sharing the text encoder parameters during training.

\subsubsection{Contrastive Learning}
Let's first introduce the training objective proposed for document triplets. In the framework of contrastive learning, we are expecting a well-trained
encoder captures meaningful similarity in the embedding space such that given an \textit{augmented} document triplet, we can get
\begin{equation*}
    similarity(\bold{a}, \bold{p}) > similarity(\bold{a}, \bold{n})
\end{equation*}
where $\bold{a}, \bold{p}, \bold{n}$ represent the semantic embedding of components in an augmented triplet.
An effective contrastive loss function, called InfoNCE \citep{oord2018representation}, is used to achieve this goal: 
\begin{equation}\label{eq: infonce}
    InfoNCE(\bold{a}, \bold{p}, \bold{n}, Z; \tau) = \frac{\exp(\bold{a}\cdot\bold{p}/\tau)}{\sum_{\bold{z}\in {Z \cup\{\bold{p}, \bold{n}\} }}\exp(\bold{a}\cdot\bold{z}/\tau)},
\end{equation}
where $\tau$ is a temperature hyper-parameter and $Z$ is the set of negative samples in the same training batch of triplets. Intuitively,
this loss is the log loss of a ($|Z|$+2)-way softmax-based classifier. In-batch negatives $Z$ is commonly used in the literature to prevent model overfitting and help future retrieval application if needed \citep{chen2020simple, feng2020language}.

\subsubsection{Multi-label Classification}
In the document-topic association dataset, each document can be mapped to multiple topics and the distribution is skewed when we consider how many documents each topic is associated with.

In this case, Binary Cross-Entropy (BCE) loss can be utilized to model multi-label setups assuming that given a document, its associations with different topics are independent. Its mathematical formulation is given below,
\begin{equation}\label{eq: bce}
BCE(d, T^d) = \sum_{t\in{T^d}}{y_t^d\log(p_t^d) + (1-y_t^d)\log(1-p_t^d)}
\end{equation}
where $d$ indicates a document and $T^{d}$ represents the set of topics positively or negatively associated with $d$.
Meanwhile, we use $y_t^d$ to represent label value for a document-topic pair. Accordingly, $p_t^d$ denotes the probability that document $d$ is predicted to be associated with $t$, computed as
\begin{equation*}
p_t^d =  \textrm{sigmoid}(\bold{w}_t \cdot \bold{d})
\end{equation*}
where $\bold{w_t}$ is a learnable parameter for topic $t$ and $\bold{d}$ is the document's [CLS] embedding.

\begin{figure}
    \includegraphics[width=.5\textwidth]{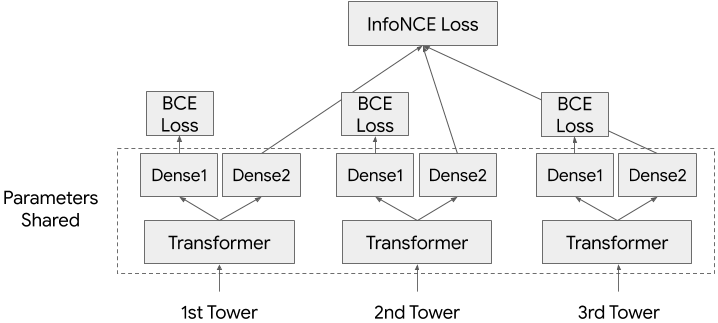}
    \vspace{-0.5cm}
    \caption{NewsEmbed model structure. }
    \label{fig: model_structure}
    \vspace{-0.3cm}
\end{figure}

\subsubsection{Model Structure}
The model structure is illustrated in Figure~\ref{fig: model_structure}. We adopt a neural structure with three towers for 
modeling triplets. Specifically, a pre-trained transformer serves as the text encoder. On top of that, [CLS] embedding from the top transformer layer is connected with two dense layers, corresponding to the semantic embedding and classification logits as described in previous paragraphs, respectively.
Similar to other triplet networks, transformer encoders and the two dense layers share parameters across towers. 

As you might notice, BCE loss is defined for single document while InfoNCE loss requires triplet input. In order to effectively co-train the two tasks together, we pack three documents in the multi-label classification dataset so that its input format is consistent with contrastive learning.

One missing details in the figure is the stop-gradient operation connected with the transformer output when the input text is after translation. As introduced at the bottom of Section~\ref{subsubsec: document_triplets}, part of the augmented triplets are augmented with translation for better language alignment.
Considering the translated text could be noisy, applying stop-gradient ensures the encoder won't be affected by possibly poor translation, while still allows the knowledge transfer from high-resource to low-resource languages.

\subsubsection{Training}
The training for NewsEmbed needs to consider both multilingual and multitask where the data scales vary by language and task. 
Similar to how multilingual BERT handles language imbalance at pre-training stage, we apply the same exponential smoothing to specify the weight for each (language, task) dataset. Its formulation is described in Appendix~\ref{subsec: tech_details} in details.

At each training step, we sample one dataset according to the dataset size and feed a random batch to the triplet network depicted in Figure~\ref{fig: model_structure}. This is also called alternate training in the literature, which turns out to be more effective than mixing all datasets together as analyzed in later experiment.



%% file: src/evaluation.tex
\begin{table}[]
\small
\begin{tabular}{r|rrrrr}
\hline

          & BERT (B) & XLM (B)  & XLM-R (L) & \textbf{NewsEmbed} \\\hline
XNLI      & 65.4  & 69.1 & \textbf{79.2} & 76.4 \\
PAWS-X    & 81.9  & 80.9 & 86.4 & \textbf{88.3} \\
UDPOS     & 71.5  & 71.3 & 73.8 & \textbf{76.5} \\
PANX & 62.2  & 61.2 & \textbf{65.4} & 62.2 \\
XQuAD-F1    & 64.5  & 59.8 & \textbf{76.6} & 74.3 \\
MLQA      & 61.4  & 48.5 & \textbf{71.6} &  70.6 \\
TyDiQA    & 59.7  & 43.6 & 65.1 &  \textbf{68.7} \\
Tatoeba   & 38.7  & 32.6 & 57.3 &  \textbf{80.1} \\\hline
Average    & 63.0 & 58.2 & 71.8 &  74.6 \\\hline

\end{tabular}
\caption{XTREME benchmarks on test dataset. We use B and L to indicate base and large model sizes.}
\label{tab: xtreme}
\vspace{-0.5cm}
\end{table}

In this section, we first show performance of our pre-trained BERT model developed from news corpus. Then we evaluate NewsEmbed, our document encoder trained with weak supervision for news, in various down-stream applications. Lastly, we conduct ablation studies to investigate alternative designs. Without specific notation, we assume the evaluated model has the same configuration as BERT-base. Reproduction details are provided in Appendix~\ref{subsec: tech_details}.

\subsection{Pre-trained Model}
As previously introduced in Section~\ref{subsec: corpus_pretraining}, our pre-training corpus comprises monolingual documents and bilingual
translation pairs. In particular, monolingual documents only includes news, which can be treated as a special slice of the web data. It is interesting to study how a model pre-trained from this dataset performs, especially when we compare it on the open-domain tasks with other models trained from the entire web without domain filtering.

We evaluate NewsEmbed with XTREME~\citep{hu2020xtreme}, a comprehensive benchmark for cross-lingual transfer learning on a diverse set of languages and tasks. Results are reported in Table~\ref{tab: xtreme}. NewsEmbed outperforms other base or slightly larger models, including multilingual BERT~\citep{devlin2019bert} and XLM~\citep{lample2019cross}. Surprisingly, as model size of NewsEmbed is the same as BERT-base, it performs highly comparable or even better than XLM-R large model~\citep{conneau2020unsupervised}. This implies that modeling news together with translation data is suitable or even better for generic natural language understanding tasks and cross-lingual transfer. We suspect data quality of news corpus together with the adoption of large translation corpus contribute to the good XTREME performance.

\begin{table}[t]
\small
\centering
\begin{tabular}{c|ccccc|c}
\hline
\textbf{Model} & \textbf{STS 12} & \textbf{STS 13} & \textbf{STS 14} & \textbf{STS 15} & \textbf{STS 16} &  \textbf{Avg.} \\\hline
SBERT & 63.8          & 69.3          & 72.9          & 75.2          & 73.3          & 71.1 \\
USE          & 65.6          & 68.0          & 71.5          &80.8          & 78.7    & 72.0 \\
mUSE         & 68.1 & 70.4 & 73.5 & \textbf{81.6} & \textbf{79.7}  & 74.2 \\
Laser        & 62.3          & 51.6          & 67.0          & 75.4          & 72.3          & 65.6 \\
LaBSE        & 66.8          & 67.7          & 69.9          & 78.8          & 74.2          & 72.1 \\
\textbf{NewsEmbed}    & \textbf{68.8} & \textbf{76.1} & \textbf{78.7} & 80.3 & 78.2 & \textbf{77.2} \\\hline
\end{tabular}
\caption{Spearman rank correlation between embedding similarity and gold labels
on STS 12-16 datasets [unsupervised]. }
\label{tab: sts_eval}
\vspace{-0.6cm}
\end{table}

\subsection{Weakly Supervised Model}
NewsEmbed is trained to be a generic document encoder for news. 
Moreover, the encoder is supposed to be robust towards text length since the training data is augmented with both short- and long-context. 
Bearing this in mind, we evaluate NewsEmbed from three perspectives, including 1) whether a task is in-domain or out-of-domain, 2) whether a task is supervised or unsupervised, and 3) whether the text is short or long.
To demonstrate the effectiveness of NewsEmbed, we compare it with the
following encoders:
\begin{itemize}
    \item Sentence-BERT (\emph{SBERT}) \citep{reimers2019sentence} \footnote{\url{https://huggingface.co/sentence-transformers/bert-base-nli-mean-tokens}}: uses transformer-based siamese and triplet network structures to derive semantic text representation. The work proposes to initialize the model from BERT pre-trained checkpoint and then trains on natural language inference corpus.
    \item Universal Sentence Encoder (\emph{USE}) \citep{cer2018universal} \footnote{\url{https://tfhub.dev/google/universal-sentence-encoder/4}}: USE trains a transformer network on web sources including news, and then augments it with SNLI dataset \citep{bowman2015large}. 
    \item multilingual Universal Sentence Encoder (\emph{mUSE}) \citep{chidambaram2019learning} \footnote{\url{https://tfhub.dev/google/universal-sentence-encoder-multilingual/3}}: mUSE extends USE to support 16 languages by training a multi-task siamese network using translation based bridge tasks.
    \item Laser \citep{artetxe2019massively} \footnote{\url{https://github.com/yannvgn/laserembeddings}}: Laser represents 93 languages via Bi-LSTM encoder trained on publicly available parallel corpora.
    \item LaBSE \citep{feng2020language} \footnote{\url{https://tfhub.dev/google/LaBSE/1}}: LaBSE embeds text sentences in a language agnostic way through training the encoder with transformer-based siamese network. The model is first pre-trained on web corpus together with translation sentence pairs, and then is fine-tuned with translation corpus alone.
\end{itemize} 

\begin{table}[t]
\small
\centering
\begin{tabular}{c|ccc}
\hline
\textbf{Model} & \textbf{News} &  \textbf{Image Caption}  &  \textbf{Forum}  \\\hline
SBERT & 72.2 & 86.7 & 64.0 \\
USE   & 67.3 & 86.8 & 67.5 \\
mUSE  & 71.7 & 88.8 & 69.0 \\
Laser & 65.0 & 76.9 & 63.9 \\
LaBSE & 75.3 & 76.3 & 62.9 \\
\textbf{NewsEmbed} & \textbf{80.9} & \textbf{88.9} & \textbf{70.8} \\\hline
\end{tabular}
\caption{Spearman rank correlation between  embedding similarity and gold labels on in-domain and out-of-domain slices of STS benchmark dataset with train, dev and test merged [unsupervised].}
\label{tab: sts_eval_domain}
\vspace{-0.5cm}
\end{table}

\subsubsection{Semantic Textual Similarity}\label{sec: sts}
Inspired by the good performance of pre-trained model in XTREME, we first evaluate the \emph{open-domain} performance on 
Semantic Textual Similarity (STS) tasks\footnote{STS dataset is collected from image captions, news headlines and user forums. Thus two out of the three genres are out-of-domain.} \citep{cer2017semeval}. Specifically, these tasks require determining how semantically similar a pair of English sentences are.

\paragraph{Unsupervised setup}
We use SentEval \citep{conneau2018senteval} tool box to evaluate models on STS tasks 2012-2016 \citep{cer2017semeval} without using any STS specific training data. These datasets provide labels between 0 and 5 as semantic relatedness of sentence pairs. Following \citet{reimers2019sentence}, we compute the Spearman's rank correlation $\rho$ between the cosine-similarity of [CLS] embeddings and the gold labels.  Performance is reported by convention as $\rho\times100$.

As shown in Table~\ref{tab: sts_eval}, although trained with news data, NewsEmbed has competitive performance comparing to existing works that are optimized for generic sentence embeddings. Since STS datasets are partially collected from news domain, we further
investigate performance variations over data domains. Results are reported in Table~\ref{tab: sts_eval_domain} for STS benchmark dataset, which is a subset of STS 12-17. One can observe that NewsEmbed outperforms other approaches by a large margin on the news slice, and is still competitive when evaluated on out-of-domain data.

\paragraph{Supervised setup}
Supervised learning is another dimension to evaluate text representation quality. The datasets in this experiment include the Microsoft Research Paraphrase Corpus (MRPC) \citep{dolan2004unsupervised}, which is a paraphrase identification dataset aiming to identify if two sentences are paraphrases of each other. The SICK Relatedness (SICK-R) dataset \citep{marelli2014sick} contain labels in the range between 1 and 5 to indicate the relatedness of two sentences. The Semantic Textual Similarity Benchmark (STS-B) \citep{cer2017semeval} follows similar label representation but the range is between 0 and 5.  On top of encoder's output, we learn a classification head for MRPC and regression head for SICK-R and STS-B using SentEval. 

Results are shown in Table~\ref{tab: sts_eval_sup}. NewsEmbed again shows strong performance as it is only slightly worse than mUSE.
The quality drop is likely because of out-of-domain dataset such as SICK-R, and the fact that NewsEmbed has multiple objectives to optimize, affecting the model capacity attributed to short-text similarity.

\begin{table}[t]
\centering
\small
\begin{tabular}{c|cccc|c}
\hline
\textbf{Model}        & \textbf{MRPC-acc}   & \textbf{MRPC-f1}   & \textbf{SICK-R} & \textbf{STS-B} & \textbf{Avg.}\\\hline
SBERT & \textbf{75.1} & 82.6 & 79.3 & 73.6  & 77.6 \\
USE          & 72.6 & 81.8 & 78.9 & 79.1 & 78.1 \\
mUSE        & 74.6 & \textbf{82.9} & \textbf{80.2} & 81.5 & \textbf{79.8} \\
Laser        & 74.1 & 82.5 & 79.0 & 77.8 & 78.4 \\
LaBSE        & 74.4 & 82.3 & 79.1 & 77.6 & 78.4 \\
\textbf{NewsEmbed}    & 73.5 & 81.9 & 79.1 & \textbf{82.3} & 79.2 \\\hline
\end{tabular}
\caption{Evaluation on semantic similarity with head trained on top of frozen embeddings [supervised]. Accuracy/F1 are used for MRPC. Spearman rank correlation is used for SICK-R/STS-B.}
\label{tab: sts_eval_sup}
\vspace{-0.7cm}
\end{table}

Based on the above experimental results, we can conclude that NewsEmbed behaves generally well at the sentence granularity regardless of the domain.

\subsubsection{News Clustering and Retrieval}
A good document encoder is supposed to work well on long contexts. We first evaluate this in an unsupervised setup, and specifically rely on document clustering and retrieval as the proxy tasks. The rational is that both these tasks depend on a good document representation.

The first dataset used in this experiment is Multi-News~\citep{fabbri2019multi}, which contains journalist-written news summaries and their corresponding  source documents. For the purpose of clustering and retrieval, we consider each summary defines a cluster and the similarity within its linked  documents should be higher than those not linked. To avoid too many small clusters, we filter the dataset on data points with at least 8 documents, which leaves us with 380 clusters and 3216 source documents.

The second dataset is Multilingual Stream News Cluster dataset (Stream-News for short) \citep{miranda2018multilingual}, proposed here for studying the multilingual performance.
It consists of news documents from English, German, and Spanish. We filter out events with size smaller than 50, yielding 11757 evaluation articles with 97 event ids.

For clustering, we apply $k$-means on document embeddings assuming the cluster size is known and report Adjusted Rand Index~\citep{rand1971objective}. For retrieval, we use Multi-News dataset with the summary as the retrieving query and report mean average precision. 

Results are summarized in Table~\ref{tab: clustering_result}.  NewsEmbed continues to perform strong when text is long. Since the data is in-domain, it outperforms all other approaches in both clustering and retrieval metrics. Besides NewsEmbed, we notice USE and mUSE are notably competitive. We suspect heavily pre-training on news data is critical as both approaches have this component. In comparison, LaBSE is pre-trained on web data but then heavily tuned on translation sentence pairs, which biases the model to excel at detecting meaning equivalence in cross-lingual environment. The rest two approaches, i.e., SBERT and Laser, lack domain adaptation for news.

\begin{table}[t]
\centering
\small

\begin{tabular}{c|cc|c}
\hline
\multirow{2}{*}{\textbf{Model}} & \multicolumn{2}{c|}{\textbf{Multi-News}} & \textbf{Stream-News}  \\\cline{2-4}
 & ARI & mAP@8 & ARI \\\hline
SBERT & 9.7 & 39.2 & 32.8 \\
USE   & 27.6 & 70.1 & 35.6 \\
mUSE  & 26.6 & 68.3 & 44.9 \\
Laser & 1.7 & 16.6 & 7.6 \\
LaBSE & 16.7 & 53.7 & 39.8 \\
\textbf{NewsEmbed} & \textbf{29.0} & \textbf{73.1}  & \textbf{51.9} \\\hline
\end{tabular}
\caption{Adjusted Rand Index for clustering on two datasets and mean Average Precision on Multi-News dataset for retrieving articles from summary [unsupervised].}
\label{tab: clustering_result}
\vspace{-0.5cm}
\end{table}

\subsubsection{News Classification}
In this experiment, we evaluate NewsEmbed in the setting of document classification.
Similar to previous experiment in Section~\ref{sec: sts}, we train a logistic regression on top of the [CLS] token embedding. We consider four popularly used datasets that cover a variety of publishers and news styles: BBC \citep{greene2006practical}, 20 Newsgroups\footnote{http://qwone.com/~jason/20Newsgroups/}, AG News~\citep{zhang2015character}, and MIND-small ~\citep{wu2020mind}, with statistics outlined in Table~\ref{tab:data_stat}.

To evaluate the effectiveness of the embeddings, we only use 10\% training data of AG News. The dataset is down-sampled because 1) logistic regression doesn't require a lot of training data, and 2) extracting embeddings is time costly for baselines. The classification accuracy scores are reported in Table~\ref{tab: classification_result}. Similar to previous unsupervised experiments, NewsEmbed continues to be the best, followed by mUSE. LaBSE ranks as the 3rd best method, and performs relatively better in supervised than unsupervised settings. This is consistent with the behaviour in the STS experiments.

\begin{table}[ht]
\small
    \centering
    \begin{tabular}{cccc}
        \hline
        \textbf{Dataset}  & \textbf{\# Classes} & \textbf{\# Train} & \textbf{\# Test}\\\hline
        BBC & 5 & 1780 & 445 \\\hline
        20 Newsgroups & 20 & 11314 & 7532 \\\hline
        10\% AG News & 4 & 12000 & 7600 \\\hline
        MIND-small & 18 & 51421 & 12856 \\\hline
    \end{tabular}
    \caption{Dataset statistics for document classification.}
    \label{tab:data_stat}
    \vspace{-0.7cm}
\end{table}
\begin{table}[h]
    \centering
    \small
    \begin{tabular}{c|cccc|c}
        \hline
        \textbf{Model}  & \textbf{BBC} & \textbf{20 Newsgroup} & \textbf{AG News} & \textbf{MIND} & \textbf{Avg.}\\\hline
        SBERT     & 96.0 & 64.5 & 86.5 & 75.2 & 80.5\\
        USE       & 97.3 & 66.4 & 85.8 & 75.0 & 81.1\\
        mUSE      & 97.3 & 71.4 & 86.8 & 75.8 & 82.8 \\
        LaBSE     & 96.6 & 70.3 & 86.3 & 73.9 & 81.8\\
        Laser     & 92.1 & 63.5 & 81.4 & 63.1 & 75.0\\
        \textbf{NewsEmbed} & \textbf{97.5} & \textbf{73.9} & \textbf{89.1} & \textbf{77.0} & \textbf{84.2}   \\\hline
    \end{tabular}
    \caption{Accuracy scores for news classification datasets with logistic classifiers on top of frozen embeddings [supervised].}
    \label{tab: classification_result}
    \vspace{-0.3cm}
\end{table}

\subsubsection{XTREME Comparison}
We compare XTREME benchmarks before and after weakly supervised learning and show the result in Table \ref{tab: xtreme_comparison}. We find that all tasks get moderate metric drop. Considering previous experiments about encoders capturing topical semantics given either short or long texts, it seems that NewsEmbed achieves outstanding performance over there via sacrificing its pre-trained reasoning and representation power at token level. 

Based on this result, it becomes interesting to co-train masked language model objective with the InfoNCE and BCE objectives in the triplet network. We leave this as future work.

\begin{table}[]
\small
\begin{tabular}{r|rr}
\hline
          & Before & After\ \\\hline
XNLI      & 76.4 & 72.8 \\
PAWS-X    & 88.3 & 84.1 \\
UDPOS     & 76.5 & 72.8 \\
PANX      & 62.2 & 61.2 \\
XQuAD     & 74.3 & 72.4 \\
MLQA      & 70.6 & 67.5 \\
TyDiQA    & 68.7 & 66.7 \\
Tatoeba   & 80.1 & 77.6 \\\hline
\end{tabular}
\caption{XTREME benchmarks on test dataset before and after weakly supervised learning.}
\label{tab: xtreme_comparison}
\vspace{-0.4cm}
\end{table}

\subsection{Ablation Studies}
\label{subsec: analysis}
\input{src/analysis.tex}

%% file: src/analysis.tex
Previous experiments are conducted upon public benchmarks, which are mostly English-only datasets. During the development of NewsEmbed, to better study alternative designs, we mainly resort to some in-house benchmarks. These include several semantic textual similarity corpora in the news domain, with the focus on evaluating sentence-to-document and document-to-document similarities. To measure model multilinguality more accurately, we record metrics in (English, Spanish, Hindi, Chinese) and compute the z-scores over specified step ranges and ablations for each benchmark. Then we aggregate all scores by taking the mean.

We mainly investigate the following ablations:
\begin{itemize}
    \item \emph{-Pre-training:} Randomly initializing the checkpoint.
    \item \emph{-Alternate training:} Not using alternate training by mixing all <language, task> datasets together.
    \item \emph{-Cross-accelerator negatives:} In-batch negatives are from the same accelerator in distributed learning.
    \item \emph{-Translation augmentation:} Removing the translation in document triplets for non-English texts.
    \item \emph{-Triplet classifier:} Removing the light-weight classifier to filter out noises in document triplets.
    \item \emph{-Document-topic associations:} Removing the document-topic associations from the datasets.
\end{itemize}
Table \ref{tab: ablation} shows the effect of removing each component from NewsEmbed model. First of all, we observe that the pre-training stage is the most critical although our weak supervision dataset is pretty large. Next, alternate training greatly improves optimization effectiveness, mainly because in-batch negatives are sharing the same languages. Similarly, increasing the effective batch size through cross-accelerator sharing also shows good improvement and noticeably brings more benefits for non-English. Our guess is that by making negative examples harder incorporates some false negatives, but it affects less for languages other than English due to smaller publisher density. Another study that gives better z-score for non-English languages is adding the translation augmentation. It verifies our conjecture about knowledge transfer by language alignment. The next ablation study shows the positive results about applying triplet classifier to clean data. It is interesting that non-English corpus improves significantly because of it, further confirming that the quality of the weak supervision is more important than the quantity. Lastly, the dataset of document-topic associations turns out to be helpful for both high- and low-resource languages, and we believe it is complementary to the document triplet dataset.

\begin{table}[t]\small
    \begin{tabular}{c|cc}
        \hline
        \multirow{2}{*}{\textbf{Ablation}}  & \multicolumn{2}{c}{\textbf{Z-score change}} \\\cline{2-3}
                            & \textbf{en} & \textbf{other languages} \\\hline
        -Pre-training & -1.35 & -1.51 \\
        -Alternate training & -0.73 & -0.97 \\
        -Cross-accelerator negatives & -0.34 & -0.67 \\
        -Translation augmentation & -0.42 & -0.72 \\
        -Triplet classifier & -0.34 & -0.70 \\
        -Document-topic associations & -0.45 & -0.46 \\
        \hline
    \end{tabular}
    \caption{Z-score changes of different ablation studies. }
    \label{tab: ablation}
    \vspace{-0.7cm}
\end{table}

%% file: src/related_work.tex


We have discussed in details in Section~\ref{sec: intro} about recent NLP developments in the area of text representation. The most related work is the recent pre-training/fine-tuning framework that significantly improve the encoder capability \citep{devlin2019bert,lample2019cross}. Here we describe the rest of related work on contrastive learning and modeling of news.


\vspace{-1mm}
\paragraph{ Contrastive learning by augmentation}
Contrastive learning can be considered as learning object representation by comparing. In computer vision, several works are proposed for self-supervised learning in a contrastive way \citep{he2020momentum, chen2020simple, misra2020self, grill2020bootstrap}. The high level idea is to encourage the model to learn representation of images so that different views or augmentations of the same image are close to the source in latent space. When it comes to textual data, \citet{feng2020language} use a dual encoder with additive margin softmax on bilingual translation pairs. \citet{fang2020cert} use back-translation as data augmentation approach that translate a sentence into a target language and back to the source language.

In our case, we mine large amount of document triplets as well as document-topic associations from the news corpus. This is based on a few observations that is unique to news domain. We consider our approach as weakly supervised because it derives noisy supervision signal based on these observations.

\vspace{-1mm}
\paragraph{News modeling}
Existing works about modeling news content mainly includes clustering, recommendation, and classification.

Works related to clustering can be categorized into online \citep{ miranda2018multilingual} and offline \citep{rehurek2010software}. They usually compute textual representation by features such as bag-of-words \citep{aggarwal2006framework}, learned topics \citep{rehurek2010software}, or deep learning representations \citep{guo2017improved}. Our work differentiates from them by using a novel approach to mine web-scale knowledge
and shows outstanding performance on clustering benchmarks.

For recommendation, a good representation of news articles and users are critical. \citet{li2011scene} represents news article using content, access patterns, named entities, popularity and recency. \citet{okura2017embedding} uses denoising autoencoder to learn embeddings based on the similarities between news articles in the same and different categories. Next, \citet{de2018news} predict next-item for users sessions by representing news based on text and metadata and modeling session-based recommendation using Recurrent Neural Networks. \citet{wu2019npa} applies a convolutional neural network (CNN) to encode news articles followed by a word-level personalized attention network. Our triplet approach is similar to \citet{okura2017embedding}, but with a much higher granularity and scale of data. We focus more on the semantic representation by only using title and body to encode the document.

News classification tasks include fake news detection \citep{oshikawa2020survey, ksieniewicz2019machine} and new category classification. People use traditional approaches such as Naive Bayes and Support Vector Machines \citep{conroy2015automatic, shu2020fakenewsnet}, and neural network approaches such as LSTM \citep{rashkin2017truth} and CNN \citep{deligiannis2018deep}. In our work, we propose to train a \emph{universal} encoder for multilingual document representation, and support fine-tuning the model on downstream classification tasks.

%% file: src/conclusions.tex
In this work, we propose to study the problem of deriving a universal document embedding within the news domain. Based on observations unique to news domain, we collect weak supervision data in large scale and in different languages. An effective neural architecture, named NewsEmbed, is then proposed to co-train contrastive learning and multi-label classification from this data. 
The model demonstrates strong performance regardless of context length through a series of unsupervised and supervised evaluations, some of which are even out-of-domain.

Nevertheless, there are a few unsolved challenges. For instance, we use the full title with truncated body from the beginning up to 512 tokens because transformer encoder is expensive with longer input, while 80\% articles have content longer than that. Some recent works have been proposed to accelerate the computation of long-text transformer with certain compromise on quality. We argue that by modeling the inductive bias where title and leading paragraphs can mostly summarize the whole document, one might achieve both efficiency and effectiveness.   Another challenge would be encoding news with modern content types including image, audio and video in a multi-modal fashion. Aligning the space among these modalities and mutually enhancing through this process will become more critical as the streaming service is becoming increasingly diverse, including but not limited to podcast, radio, short video, live broadcast, etc.
Lastly, our approach mostly relies on time-sensitivity to remove noise and therefore lacks training signal to model evolving topics in long running stories. We are actively exploring those directions as future work.

%% file: src/appendix.tex
\newpage

\section{Short Text Packing at Pre-training }
During data preparation, we observe that some documents and translation sentences are much shorter comparing to standard BERT input sequence length 512. This can cause lots of paddings in order to run on TPU, making the pre-training inefficient. To fully use sequence length, we apply a greedy  algorithm to pack multiple short sequences into one training example. The packing algorithm achieves 8.5 compression ratio for the dataset of translation sentence pairs.

As shown in Algorithm~\ref{alg: packer}, a priority instance packer has attributes as cache capacity C, max sequence length L, and min packing proportion $\rho$. The packer maintains a map \textbf{M} of partially packed instances with key as sequence length (with tie-breaker by instance id) ordered non-increasingly. In practice, we use Apache Beam to process the large scale data.

\begin{algorithm}[h]
\SetAlgoLined
 \KwData{$N$ Training Instances $\{X_i\}_{i=1}^N$}
 \KwResult{List of Packed Training Instances $\textbf{Y}$}
 \textbf{Y} = []\;
 \For {$i$ in 1:N}{
    processed = False\;
    \For{$k$, $v$ in \textbf{M}}{
        \If{len($X_i$) + len(v) > L}{
            continue;
        }
        remove $k$ from \textbf{M}\;
        $\tilde{v}$ = instance by concatenating $v$ and $X_i$ \;
        processed = True\;
        \eIf{len($\tilde{v}$) $\ge \rho$L}{
            $\textbf{Y}$.append($\tilde{v}$)\;
        }{
            \textbf{M}[(len($\tilde{v}$), $i$)] = $\tilde{v}$\;
        }
        break;
    }
    \If{\text{not processed}} {
        \textbf{M}[(len($X_i$), $i$)] = $X_i$\;
        \If{len(\textbf{M}) > C}{
            remove the first element $(k_f, v_f)$ from \textbf{M}\;
            $\textbf{Y}$.append($v_f$)\;
        }
    }
 }
 \caption{Training instance packing algorithm}
 \label{alg: packer}
\end{algorithm}

\section{Technical Details in Weakly Supervised Learning}
\label{subsec: tech_details}
\subsection{Training}
We initialize NewsEmbed from a BERT-style pre-trained checkpoint described in Section~\ref{subsec: pretrain} and train triplet network with weak supervision using 256 TPU v3 chips. The batch size is 8,196 and we train the model for 400k steps, which is roughly three epochs. Similar to BERT, Adam optimizer is deployed with learning rate set as 5e-5.

In the contrastive learning objective, i.e., Equation~\ref{eq: infonce}, temperature $\tau$  is set as 0.05 and the vectors in the augmented triplet are normalized by their L2-norm before computing the inner product. In-batch negatives $Z$ include cross-accelerator negatives to allow us to fully realize the benefits of distributed training.

In the multi-label classification objective, i.e., Equation~\ref{eq: bce}, $T$ comprises of 600 hand-picked topics. Positive to negative ratio is empirically set as 25\% through down-sampling mentioned in Section~\ref{subsec: document-topic}.

During the training, the total loss is computed as the summation of Equation~\ref{eq: infonce} and Equation~\ref{eq: bce} without weights.

\subsection{Languages and vocabulary}
We select a total of 101 languages that have footprints in our proprietary system.
To ensure a smoother distribution so that the model can be later trained better on low resource data, we use an exponential up-sampling approach \citep{lample2019cross} with smooth coefficient 0.7. More concretely, suppose the $N$ languages have counts distributed as $\{n_i\}_{i=1...N}$ with $n_1 \ge n_2 \ge \cdots \ge n_N$, we will up-sample the articles so that after sampling, language $i$ has expected $m_i$ articles, where
$$
m_i = n_1\left(\frac{p_i}{p_1}\right)^\alpha.
$$
Here $\alpha$ is the smooth coefficient between 0 and 1 inclusive. If $\alpha = 0$, every language is re-sampled to the largest language. If $\alpha = 1$, the distribution will remain unchanged. If $m_i$ is not an integer, we draw a random Bernoulli variable with probability of $m_i$'s fractional part to round it to an integer.

We use a large vocabulary size of 500K and apply wordpiece algorithm to learn the tokens from the above smoothed dataset. Note that the vocabulary is large because multilingual
transformer models can benefit from allocating a higher proportion of the total number of parameters to the
embedding layer \cite{conneau2020unsupervised} while the time complexity of embedding lookup operation is almost constant w.r.t. vocabulary size.

\section{Generalization Capability}
It is challenging to model news partially because of the out-of-vocabulary concepts and evolving events.
A good encoder not only needs to fit existing data well, but also is supposed to show outstanding generalization capability for future content.

Realizing this, we design the following experiment to measure performance degeneration over time. Firstly, we ensure our evaluation benchmarks only contain recently published documents after January 1, 2020. Next, two training datasets are re-collected by deleting documents published after January 1, 2020 and January 1, 2019 respectively. After that, we retrain NewsEmbed on them and compare their evaluation metrics with the model trained on the complete corpus (containing documents up to June 30, 2020). We consider the differences in benchmark metrics should reflect our system's generalization capability.

The reason we choose January 1, 2020 as the partition date is because of the overwhelming discussions about COVID-19 on the web regardless of languages. By training a model on datasets without any COVID-related topics, we expect this study can measure the model's staleness more accurately and guide us how often to re-train a model to retain reasonable performance.

It is worth noting that NewsEmbed is initialized on a pre-trained checkpoint. To prevent it from leaking information, we apply the same data filter at the pre-training stage.

Results are reported in Table~\ref{tab: generalization}. As one can tell, z-score of the model trained on fresher training dataset drops less, verifying that the model quality becomes worse over time. Moreover, due to the topic-drift brought by COVID-19, the loss becomes significantly large in the recent six months, regardless of the languages.
This guides us to retrain the model every few months especially when there are major breaking events happening.

\begin{table}[h]
\small
\vspace{-0.2cm}
    \begin{tabular}{c|cc}
        \hline
        \multirow{2}{*}{\textbf{Ablation}}  & \multicolumn{2}{c}{\textbf{Z-score change}} \\\cline{2-3}
                            & \textbf{en} & \textbf{other languages} \\\hline
        Baseline (up to 06/30/2020) & 0.00 & 0.00 \\
        - 6 months (up to 01/01/2020) & -1.74 & -0.79\\
        - 18 months (up to 01/01/2019) & -2.16 & -1.27 \\
        \hline
    \end{tabular}
    \caption{Z-score changes after removing recent documents from training datasets.  }
    \label{tab: generalization}
    \vspace{-0.5cm}
\end{table}